\documentclass[]{rtp_tech_report}

\usepackage{arydshln}        
\usepackage{float}
\usepackage{wrapfig}
\usepackage{textcomp}
\usepackage{colortbl}
\usepackage{fixmath}
\usepackage{algorithmicx,algorithm,algpseudocode}
\usepackage{soul}
\usepackage{diagbox}
\usepackage{comment}
\usepackage{afterpage}
\usepackage{nicefrac}

\newcommand{\name}{{\textsc{RTPurbo}}\xspace}

\definecolor{L0}{RGB}{0,0,0}
\definecolor{L1}{RGB}{210,2,0}
\definecolor{L2}{RGB}{245,197,3}
\definecolor{L3}{RGB}{249,153,53}
\definecolor{L4}{RGB}{0,160,234}
\definecolor{L5}{RGB}{37,172,116}
\definecolor{L6}{RGB}{139,9,135}
\definecolor{L7}{RGB}{215,160,80}
\definecolor{L8}{RGB}{229,223,135}
\definecolor{L9}{RGB}{130,140,71}
\definecolor{L10}{RGB}{186,101,81}
\definecolor{L12}{RGB}{25,99,164}
\definecolor{L14}{RGB}{178,137,195}
\definecolor{L17}{RGB}{12,68,85}

\newcommand{\SOTA}[1]{\textbf{#1}}

\newcolumntype{C}{>{\centering\arraybackslash}X}

\title{Full Attention Strikes Back: Transferring Full Attention into Sparse within Hundred Training Steps}

\author{%
\parbox{0.9\textwidth}{\centering
Yanke Zhou$^{1,\sharp}$\hspace{1.0em}%
Yiduo Li$^{2}$\hspace{1.0em}%
Hanlin Tang$^{2,\dagger}$\hspace{1.0em}%
Maohua Li$^{1,\sharp}$\hspace{1.0em}%
Kan Liu$^{2}$\\[0.45em]
Tao Lan$^{2}$\hspace{1.0em}%
Lin Qu$^{2}$\hspace{1.0em}%
Yuan Yao$^{1,\S}$\hspace{1.0em}%
Xiaoxing Ma$^{1}$
}
}

\affiliation{%
\parbox{\textwidth}{\centering\small
$^1$Nanjing University \quad $^2$Alibaba Group
}
}

\renewcommand{\contributionlist}{\vspace{-4mm}}

\newcommand{\authorfootnotes}{%
  \begingroup
  \renewcommand{\thefootnote}{}%
  \footnotetext{\textsuperscript{\ensuremath{\dagger}}Project lead \quad \textsuperscript{\S}Corresponding author \quad \textsuperscript{\ensuremath{\sharp}}Work done during internship at Alibaba}%
  \endgroup
}

\checkdata[E-mail]{\email{yankezhou@smail.nju.edu.cn wanghaisheng.whs@alibaba-inc.com y.yao@nju.edu.cn}}

\abstract{
Long-context inference in large language models is bottlenecked by the quadratic cost of full attention. Existing efficient alternatives often rely either on native sparse training or on heuristic token eviction, creating an undesirable trade-off among efficiency, training cost, and accuracy. In this work, we show that full-attention LLMs are already intrinsically sparse and can be transformed into highly sparse models with only minimal adaptation. Our approach is built on three observations: (1) only a small subset of attention heads truly requires full long-context processing; (2) long-range retrieval is governed primarily by a low-dimensional subspace, allowing relevant tokens to be retrieved efficiently with a 16-dimensional indexer; and (3) the useful token budget is strongly query-dependent, making dynamic top-$p$ selection more suitable than fixed top-$k$ sparsification. Based on these insights, we propose \name, which retains the full KV cache only for retrieval heads and introduces a lightweight token indexer for sparse attention. By exploiting the model's intrinsic sparsity, \name achieves sparsification with only a few hundred training steps. Experiments on long-context benchmarks and reasoning tasks show that \name preserves near-lossless accuracy while delivering substantial efficiency gains, including up to a 9.36$\times$ prefill speedup at 1M context and about a 2.01$\times$ decode speedup. These results suggest that strong sparse inference can be obtained from standard full-attention training without expensive native sparse pretraining.
}

\begin{document}
\raggedbottom

\maketitle
\authorfootnotes

\begin{figure}[!htbp]
    \centering
    \includegraphics[width=\linewidth]{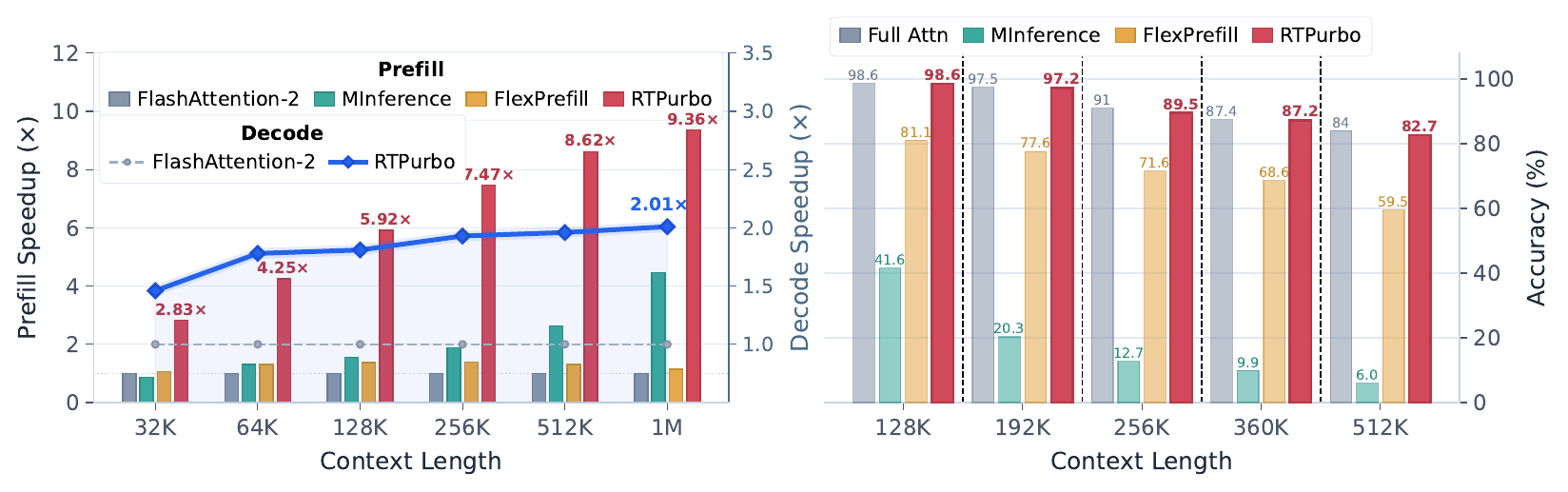}
    \caption{A teaser view of the efficiency and accuracy gains of \name.}
    \label{fig:overall}
\end{figure}

\section{Introduction}

Long-context capability has become a core requirement for modern large language models (LLMs), especially for applications such as multi-turn dialogue, long-horizon reasoning, and document understanding~\citep{deepseekR1,kimi2,qwen1M,gemini25}. However, the cost of full attention grows rapidly with context length, making long-context inference a major efficiency bottleneck. Sparse attention thus emerges as a natural direction for reducing inference cost~\citep{streamLLM,spargeattn,zucchet2026the}. 


Although many recent advances in this area replace standard full attention with more efficient alternatives, such as Kimi Delta Attention~\citep{kimiteam2025kimilinearexpressiveefficient} and DeepSeek Sparse Attention~\citep{dsa}, our study suggests that models trained with full attention already exhibit substantial intrinsic sparsity. Prior work has partially revealed this phenomenon. Specifically, sparsity arises at both the head level and the token level: most heads rely primarily on local information~\citep{streamingLLM, razorattn,duoattn}, whereas for each query only a small subset of tokens receives substantial attention mass~\citep{fasa,Quest,snapkv}. This observation naturally raises a key question:
\textit{What is the minimal surgery required to transform a full-attention model into a highly sparse one while preserving its capabilities?}

We identify three challenges:
\begin{itemize}[leftmargin=*, itemindent=0pt, itemsep=3pt, topsep=2pt, parsep=0pt, partopsep=0pt]
\item Head selection: a robust metric is needed to identify the heads that genuinely require full-context access.
\item Efficient token indexing: a lightweight selector is needed to identify the necessary tokens efficiently.
\item Adaptive sparsity: because different queries require different numbers of attended tokens, a static sparsity budget can lead to information loss.
\end{itemize}

Our method, \name, is designed to address these challenges with minimal adaptation.
The design of \name is grounded in both LLM interpretability and theoretical analysis. Prior work on inductive heads shows that some heads implement a retrieval mechanism by attending to previously similar tokens~\citep{olsson2022incontextlearninginductionheads}. Follow-up work further shows that, in long-context settings, these heads are primarily responsible for remote retrieval, whereas the remaining heads focus on local context~\citep{razorattn}. This observation motivates our head-wise design: we retain the full KV cache only for retrieval heads and discard remote tokens for local heads.

For retrieval heads, the key challenge is to identify relevant tokens efficiently. Our analysis shows that high-frequency components contribute little to long-range retrieval and can even interfere with it, suggesting that the retrieval process is governed largely by a low-dimensional subspace. This hypothesis is strongly supported by experiments: with our trained low-dimensional projector, we achieve over $90\%$ recall using only 16 dimensions. Moreover, our analysis suggests that a static Top-$k$ selector can fail in certain cases, whereas a Top-$p$ selector better adapts to the attention distribution and yields substantially better accuracy on both reasoning and long-context tasks.

Finally, we find that self-distillation is particularly effective for recovering the performance of the sparsified model. Aligning the sparse model's outputs with those of the original model substantially reduces the risk of overfitting, and only a few hundred training steps (about 1M label tokens) are required for this alignment stage. This result further supports our claim that \name performs only minimal surgery on the original model.

To the best of our knowledge, \name is the first method to achieve such near-lossless compression with lightweight continual training. Coupled with our custom sparse kernels, \name delivers up to a 9.36$\times$ speedup in prefill and a 2.01$\times$ speedup in decoding (Figure~\ref{fig:overall}). Importantly, the sparsification paradigm of \name remains highly interpretable. More broadly, our results highlight an overlooked point for full-attention models: \textit{even without native sparse training, a fully trained model can be sparsified with very small additional cost while preserving strong performance}. This finding suggests that full-attention training remains a highly competitive and practical choice.

\section{Insight Behind \name}

\begin{figure}[!t]
    \centering
    \includegraphics[width=0.6\linewidth]{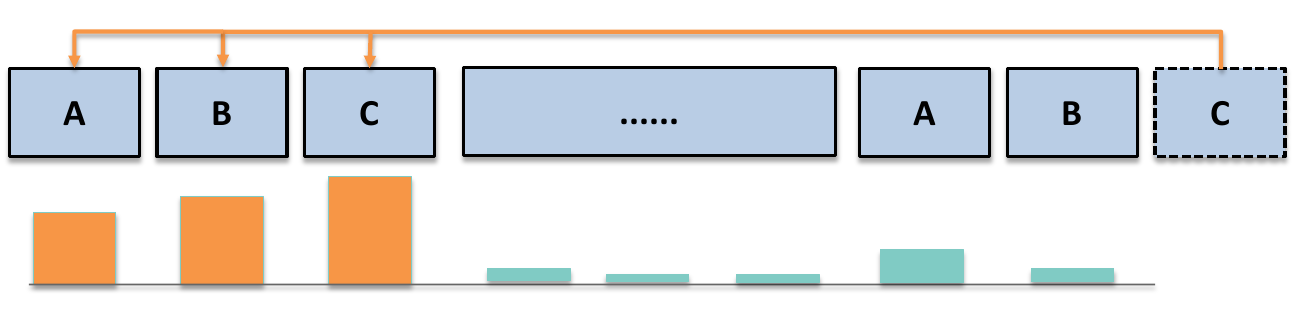}
    \caption{Unlike most attention heads that mainly focus on local information, retrieval heads attend to regions that are semantically related to the current query token (e.g. similar pattern), even when those regions are far away in the context.}
    \label{fig:retrieval_head_show}
\end{figure}

\subsection{Head Specialization as a Natural Prior for Sparse Attention}

Recent studies suggest that attention heads in pretrained LLMs are not homogeneous, but instead specialize into distinct functional roles. In particular, prior work has shown that only a small subset of heads is responsible for retrieving distant relevant content, while many others mainly process local information~\citep{duoattn,razorattn}. We refer to this subset as \emph{retrieval heads}. Their characteristic behavior is to place strong attention on earlier context surrounding semantically related content, thereby exhibiting an information-retrieval pattern, as illustrated in Figure~\ref{fig:retrieval_head_show}. 

This observation provides an important design motivation for our method: \emph{we can naturally exploit the sparsity structure that the model has already formed.} Concretely, we retain the full KV cache only for retrieval heads, while for the remaining heads, which are already intrinsically sparse, we can safely discard remote tokens.

\subsection{RoPE Induces a Compressible Geometry for Retrieval Heads}
\label{sec:motiv_lowdim_retrieval}

Retrieval heads should assign high attention to semantically related tokens even when they are far apart.


However, this property of retrieval heads appears, at first glance, to be in tension with RoPE~\citep{rope}. For a query token at position $m$ and a key token at position $n$ with dimension $d = 2D$, RoPE injects position through a rotation matrix:
{\setlength{\abovedisplayskip}{4pt}\setlength{\belowdisplayskip}{4pt}\setlength{\abovedisplayshortskip}{2pt}\setlength{\belowdisplayshortskip}{2pt}
\begin{equation}
\label{eq:rope_pair_rotation}
R_i(m)=
\begin{pmatrix}
\cos(m\theta_i) & -\sin(m\theta_i)\\
\sin(m\theta_i) & \cos(m\theta_i)
\end{pmatrix},
\qquad
q_m=R(m)q,\; k_n=R(n)k,
\end{equation}}
where $R(m)=\mathrm{diag}(R_1(m),\dots,R_D(m))$, and $\theta_i$ decreases with the channel index. The resulting query--key score depends only on the relative offset $\Delta=m-n$:
{\setlength{\abovedisplayskip}{4pt}\setlength{\belowdisplayskip}{4pt}\setlength{\abovedisplayshortskip}{2pt}\setlength{\belowdisplayshortskip}{2pt}
\begin{equation}
\label{eq:rope_relative_rewrite}
s(m,n)=q_m^{\top}k_n
=\sum_{i=1}^{D}\left[a_i(q,k)\cos(\theta_i\Delta) + b_i(q,k)\sin(\theta_i\Delta)\right],
\end{equation}}
where $a_i$ and $b_i$ are bilinear coefficients induced by the $i$-th rotary pair. Equation~\eqref{eq:rope_relative_rewrite} reveals the key distinction directly: high-frequency components vary rapidly with $\Delta$ and become distance-sensitive at long range, whereas low-frequency components change smoothly and better preserve retrieval signals. This leads to our second core insight: \emph{we can reconstruct retrieval-head attention in a much lower-dimensional space.}


We therefore use this low-frequency structure as a compact retrieval subspace, enabling low-cost token selection without full-dimensional scoring.

\subsection{Retrieval Heads Require Dynamic Thresholding}

\begin{figure}[!t]
    \centering
    \begin{subfigure}[t]{0.499\linewidth}
        \centering
        \includegraphics[width=\linewidth]{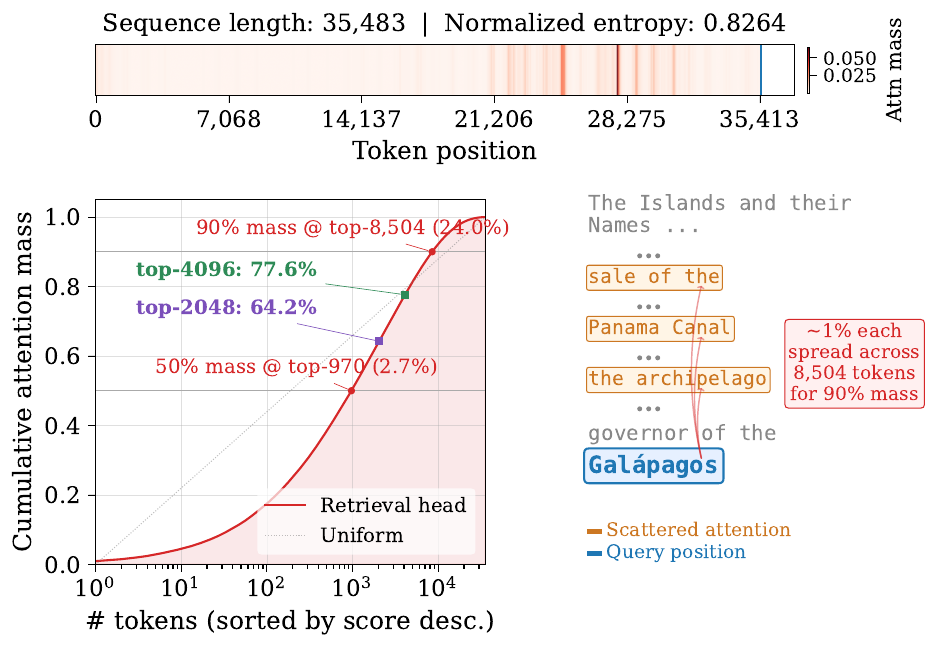}
        \caption{Diffuse retrieval triggered by the query token ``Gal\'apagos'' in a long passage.}
        \label{fig:diffuse_attn}
    \end{subfigure}\hfill
    \begin{subfigure}[t]{0.499\linewidth}
        \centering
        \includegraphics[width=\linewidth]{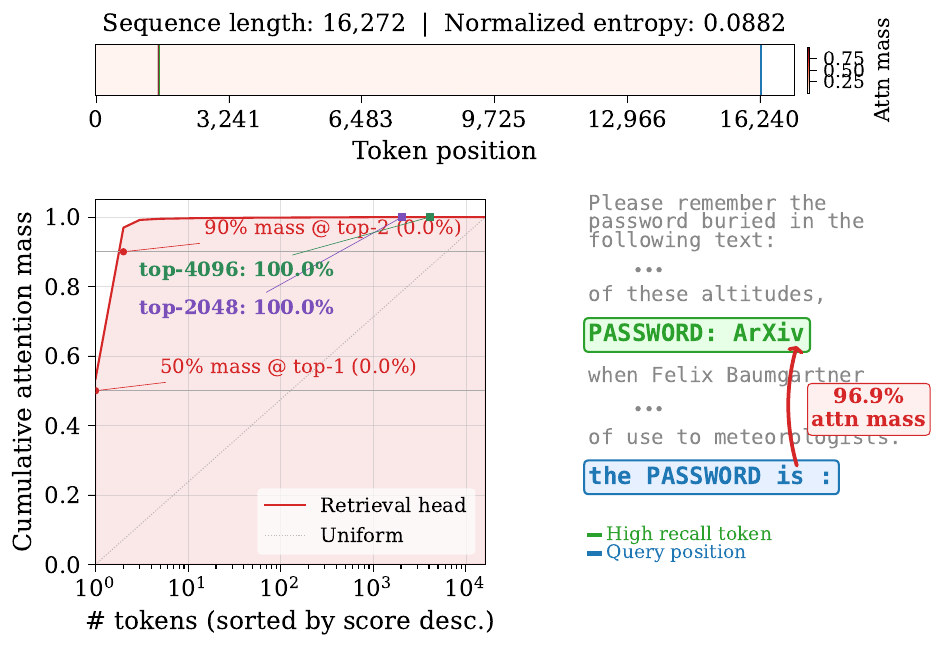}
        \caption{Concentrated retrieval in a NIAH query.}
        \label{fig:mass_attn}
    \end{subfigure}
    \caption{Retrieval-head behavior is strongly query-dependent. (a) The query token ``Gal\'apagos'' induces diffuse retrieval over many semantically related earlier tokens: about 8k tokens are needed to recover 90\%+ attention mass, while top-4k recovers only about 75\%. (b) For a needle-in-a-haystack query, retrieval is highly concentrated: two tokens recover 96.6\% attention mass, whereas top-4k retains many unnecessary tokens.}
    \label{fig:query_dependent_patterns}
\end{figure}

The remaining question is how many tokens a retrieval head should preserve once relevance can be estimated efficiently. Our findings suggest that this quantity is fundamentally query-dependent. Even within the same retrieval head, different inputs can induce very different patterns: some queries trigger broad retrieval over many distant tokens, while others lock onto only a few key tokens. The required sparsity level is therefore not a fixed attribute of the head; it changes with the query.


Figure~\ref{fig:query_dependent_patterns} illustrates this point. In one case, the query activates a broad semantic field, so the retrieval head must preserve a wide support to recover most of the attention mass. In another, the query only needs to recover a single key fact, so the head is naturally highly concentrated. 

\begin{wraptable}[10]{r}{0.40\linewidth}
    \vspace{-1.2em}
    \centering
    \scriptsize
    \setlength{\tabcolsep}{2.2pt}
    \renewcommand{\arraystretch}{0.94}
    \captionsetup{width=\linewidth, justification=raggedright, singlelinecheck=false, skip=2pt}
    \caption{Fixed top-$k$ trades recall for sparsity: top-16k computes about 8k extra tokens than top-$p$, but recovers only 3.8\% more attention mass.}
    \label{tab:topk_topp_inline}
    \begin{tabular*}{\linewidth}{@{\extracolsep{\fill}}lcc@{}}
        \toprule
        \textbf{Method} & \textbf{Tokens Recall} & \textbf{Attn Mass} \\
        \midrule
        top-2k & 2048 & 64.2\% \\
        top-4k & 4096 & 77.6\% \\
        top-16k & 16384 & 93.8\% \\
        top-$p$ 0.9 & 8504 & 90.0\% \\
        Full Attn & 35483 & 100\% \\
        \bottomrule
    \end{tabular*}
    \vspace{-2.0em}
\end{wraptable}

This is exactly where fixed-budget rules such as top-$k$ sampling become problematic. When $k$ is too small, diffuse queries recover too little attention mass and the approximation becomes inaccurate. When $k$ is too large, the retained set is no longer sparse enough and much of the extra computation is wasted. Table~\ref{tab:topk_topp_inline} makes this trade-off concrete: top-16k recovers only 3.8\% more attention mass than dynamic top-$p$, but requires computing about 8k additional tokens. The issue is therefore not choosing a better global $k$; any fixed $k$ is mismatched to the query-dependent nature of retrieval heads.



\section{Method}

We introduce \name, a head-wise attention framework with precise token-level sparse computation. This section is organized as follows. We first describe how to identify retrieval heads through offline calibration in Section~\ref{sec:offline_calibration}. We then present our sparse computation pattern in Section~\ref{sec:adaptive_sparse_attention}. Next, we describe the two-stage training pipeline required by \name in Section~\ref{sec:two_stage_training}. Finally, we describe the hardware-aware decoding kernel in Section~\ref{sec:fast_top_p_kernel}.

\subsection{Offline Head-wise Calibration}
\label{sec:offline_calibration}
To identify retrieval heads, we construct a lightweight calibration sequence by inserting an identical ``needle'' span at both the beginning and the end of a long document sampled from FineWeb~\citep{fineweb}. We quantify a head's retrieval capability by measuring the attention mass directed from the later needle to the earlier one. Let $\mathcal{N}_{\mathrm{pre}}$ and $\mathcal{N}_{\mathrm{post}}$ denote the token indices of the earlier and later needle spans, respectively. The retrieval score for head $h$ is compactly defined as:
\begin{equation}
\label{eq:head_retrieval_score}
R_h = \frac{1}{|\mathcal{N}_{\mathrm{post}}|}\sum_{t\in\mathcal{N}_{\mathrm{post}}} \sum_{j\in\mathcal{N}_{\mathrm{pre}}} A_h(t,j),
\end{equation}
where $A_h(t,j)$ represents the normalized attention score (i.e., post-softmax) from token $t$ to token $j$. 

The head retrieval behavior is highly stable and largely input-agnostic. 
Therefore, in practice, running this calibration on just one single long text sequence is sufficient to robustly score and partition all query heads into a retrieval set $\mathcal{H}_{\mathrm{ret}}$ (top-scoring heads) and a local set $\mathcal{H}_{\mathrm{loc}}$. This partition process is executed only once offline.

\begin{figure}[!t]
    \centering
    \includegraphics[width=\linewidth]{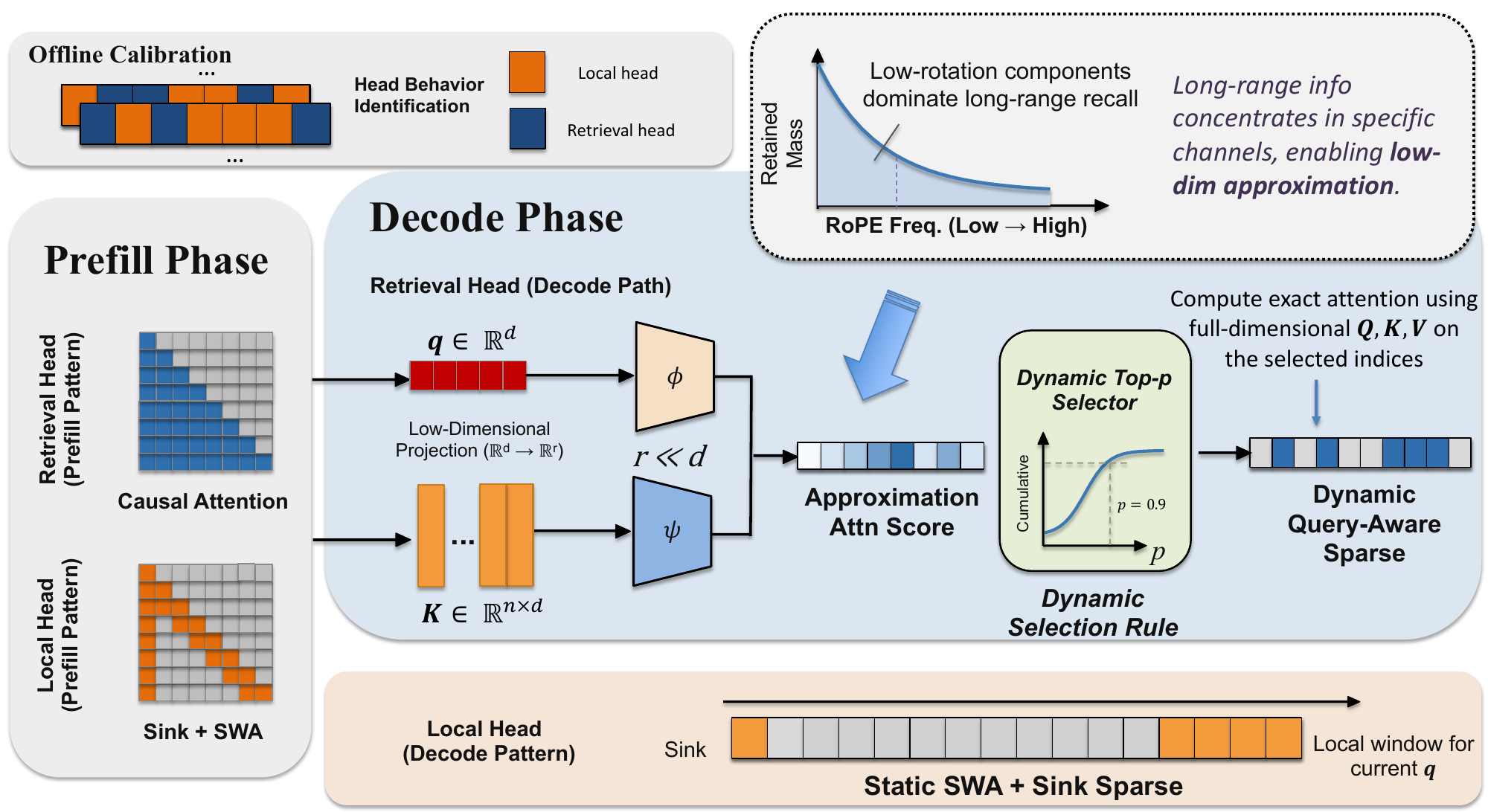}
    \caption{Overall architecture of \name.}
    \label{fig:architecture}
\end{figure}

\subsection{Adaptive Sparse Attention Mechanism}\label{sec:adaptive_sparse_attention}
During inference, local heads $h \in \mathcal{H}_{\mathrm{loc}}$ consistently apply a sliding window with attention sinks~\citep{streamingLLM} across both prefill and decode stages. In contrast, retrieval heads $h \in \mathcal{H}_{\mathrm{ret}}$ perform full dense attention during prefill to build the complete KV cache, but switch to a query-aware dynamic sparse selection during decoding. As analyzed in Section~\ref{sec:motiv_lowdim_retrieval}, high-frequency RoPE components degrade long-range affinity. To circumvent this, we estimate query-key relevance using low-rank projections $W^Q_h, W^K_h \in \mathbb{R}^{r \times d_h}$ ($r \ll d_h$) applied to the features \textbf{before} RoPE injection:
\begin{equation}
    s_h(m,n) = (W^Q_h q_{m,h}^{\mathrm{pre}})^{\top} (W^K_h k_{n,h}^{\mathrm{pre}}),
\end{equation}
where $q_{m,h}^{\mathrm{pre}}$ and $k_{n,h}^{\mathrm{pre}}$ are the pre-RoPE representations. We then construct a dynamic active set from the projected scores and compute sparse attention as
\begin{equation}
O_h(m) = \sum_{n \in \mathcal{S}_h(m)} \frac{\exp(q_{m,h}^{\top} k_{n,h} / \sqrt{d_h})}{\sum_{j \in \mathcal{S}_h(m)} \exp(q_{m,h}^{\top} k_{j,h} / \sqrt{d_h})} v_{n,h}, \qquad \mathcal{S}_h(m) = \operatorname{Top\text{-}P}\!\left(s_h(m,\cdot), p\right).
\end{equation}
In this way, the low-rank pre-RoPE projections serve strictly as an efficient routing mechanism, while the final token generation preserves the complete feature space and exact relative positional geometry. For MQA and GQA models, the resulting sparsity should be interpreted from two perspectives because our head partition is defined over query heads. \emph{Compute sparsity} is measured at the query-head level and can be viewed as the average number of attended tokens over heads. \emph{Memory sparsity} is measured at the KV-head level: for each KV head, the actual retained set is the union of the token sets selected by all query heads mapped to that KV head.

\subsection{Low-cost Two-Stage Training}
\label{sec:two_stage_training}

We adopt a lightweight two-stage training pipeline to fully restore model capabilities under the sparse regime.
In the first stage, we keep the backbone LLM frozen and independently train the low-dimension projection weights $W^Q_h, W^K_h$ for each retrieval head $h \in \mathcal{H}_{\mathrm{ret}}$. Let $a_h^{\mathrm{full}}(m)$ be the original exact attention distribution and $a_h^{\mathrm{proj}}(m; W^Q_h, W^K_h)$ be the distribution derived from the low-dimensional projected scores. We optimize the projections by minimizing the Kullback-Leibler (KL) divergence between them:
\begin{equation}
    \mathcal{L}_{\mathrm{proj}} = \sum_{h \in \mathcal{H}_{\mathrm{ret}}} \mathrm{KL}\!\left(a_h^{\mathrm{full}}(m) \,\|\, a_h^{\mathrm{proj}}(m; W^Q_h, W^K_h)\right).
\end{equation}

In the second stage, we insert the trained projections, switch to the sparse attention mode, and perform end-to-end self-distillation. The sparse model acts as a student learning to match the dense teacher's next-token predictions. Crucially, compared to standard supervised fine-tuning, \emph{self-distillation bypasses the negative impact of specific dataset distributions, thereby eliminating the tedious need to ablate and tune data mixtures.} To further reduce computational overhead, we align only the top-10 logits of the teacher. Letting $z^{\mathrm{dense}}_{(10)}$ and $z^{\mathrm{sparse}}_{(10)}$ denote the respective logits restricted to these top-10 entries, we minimize:
\begin{equation}
\mathcal{L}_{\mathrm{distill}} = \mathrm{KL}\!\left(\mathrm{softmax}(z^{\mathrm{dense}}_{(10)})\,\|\,\mathrm{softmax}(z^{\mathrm{sparse}}_{(10)})\right).
\end{equation}

\subsection{Hardware-Aware Fast Top-\texorpdfstring{$p$}{p} Decoding Kernel}\label{sec:fast_top_p_kernel}

\begin{figure}[t]
    \centering
    \includegraphics[width=\linewidth]{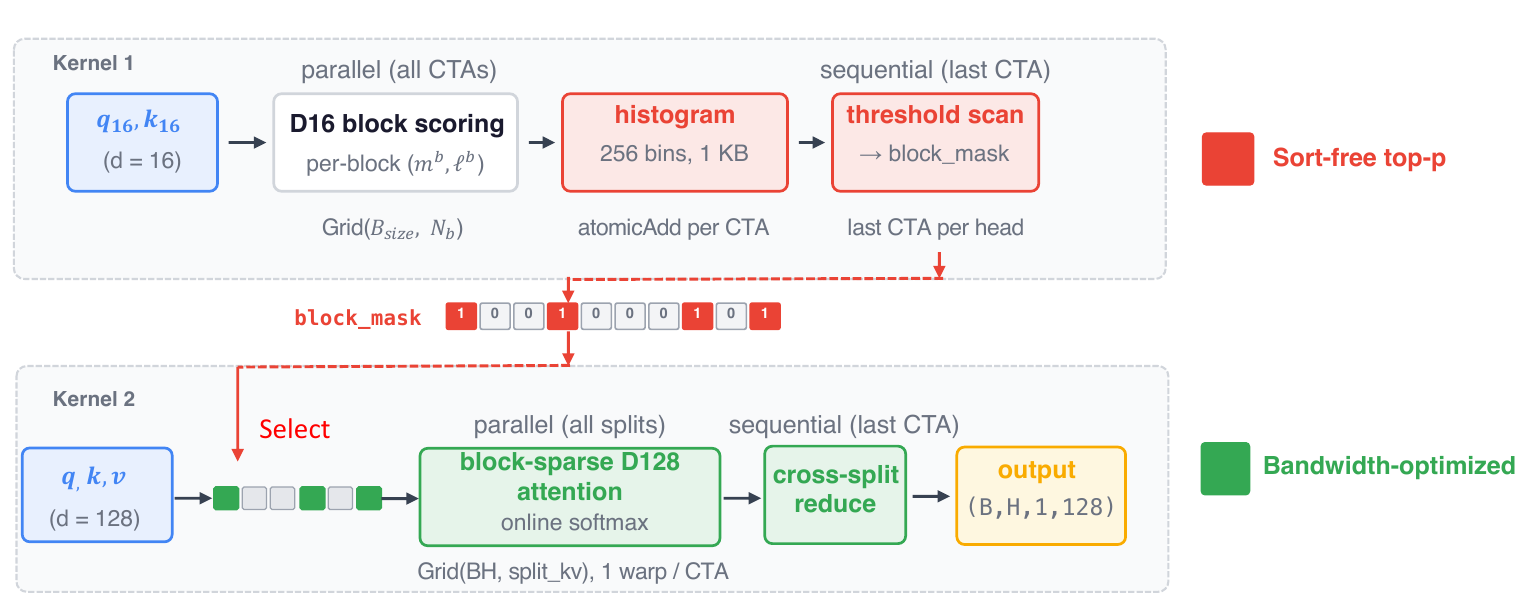}
    \caption{Overview of the hardware-aware decoding kernel in \name.}
    \label{fig:kernel_overview}
\end{figure}

We implement the block-wise top-$p$ sparse decoding using a custom GPU kernel that addresses two primary engineering challenges: (1) fast top-$p$ thresholding without expensive sorting, and (2) memory-efficient sparse decoding over long contexts.

\noindent\textbf{Sort-free top-\texorpdfstring{$p$}{p} via histogram.} 
We partition compressed K sequence into $N_b$ blocks, where each CTA (Compute Thread Array) computes a low-dimensional attention score for one block and reduces it to a block-level log-sum-exp pair $(m_b, \ell_b)$.
Since commonly used fast sorting methods still incur $O(N_b \log N_b)$ complexity, while binary-search selection requires $O(N_b)$ memory per head, which becomes prohibitive at long context where $N_b$ can exceed $16\text{K}$, we instead have each CTA atomically deposit $\ell_b$ into a 256-bin histogram indexed by $m_b$, which requires only 1 KB per head regardless of sequence length.
To avoid an additional kernel launch for the selection phase, each CTA atomically increments a per-head counter upon completion, and the last CTA to finish proceeds to scan the histogram from the highest bin, identifies the score threshold at which the cumulative attention mass reaches $p_\text{top}$, and writes a block-level binary mask.
This fuses scoring and selection into a single kernel launch with $O(1)$ memory overhead.

\noindent\textbf{Bandwidth-optimized sparse decoding.} 
For long sequences, even sparse attention remains memory-bound because the selected KV blocks can still span tens of thousands of tokens. We address this by designing a single-warp CTA with no shared memory, which keeps all state in registers and allows the SM to maximize concurrent CTAs and thus outstanding memory requests.
The inner loop is 2-token unrolled, issuing all K and V loads upfront via vectorized $\texttt{half2}$ instructions so that the subsequent score computation and online-softmax update overlap with in-flight loads.
When $B{\times}H$ alone is insufficient to fill the GPU, we further partition the KV range of each head into multiple splits, each handled by a separate CTA, and fuse the cross-split reduction into the last completing CTA via the same atomic-counter technique.

\section{Experiments}

All experiments are conducted on NVIDIA H20 GPUs with Python 3.14, CUDA 12.8, and PyTorch 2.8. For accuracy evaluation, we use the \texttt{lm-eval} framework~\citep{eval-harness} as the unified evaluation pipeline. 

\subsection{Accuracy Evaluation}
\label{sec:accuracy_eval}

\noindent\textbf{Benchmarks and Models.} We evaluate \name on two categories of benchmarks. The first category consists of long-context benchmarks, including LongBench~\citep{bai-etal-2024-longbench} and RULER~\citep{hsieh2024ruler}, which evaluate overall long-context processing ability. The second category consists of reasoning benchmarks, including AIME24~\citep{AIME24}, AIME25~\citep{AIME25}, and MMLU-PRO~\citep{mmlupro}, which are used to assess both long-decode performance and the general reasoning ability of the sparsified model. For the first category, we use Qwen3-Coder-30B-A3B, and for the second category, we use Qwen3-30B-A3B-Think, a reasoning-specialized model~\citep{qwen3}.

\begin{wraptable}[8]{r}{0.30\linewidth}
    \vspace{-1.0em}
    \centering
    \scriptsize
    \setlength{\tabcolsep}{3pt}
    \renewcommand{\arraystretch}{0.95}
    \captionsetup{width=\linewidth, justification=centering, singlelinecheck=true, skip=2pt}
    \caption{\name config}
    \label{tab:exp_settings}
    \begin{tabular*}{\linewidth}{@{\extracolsep{\fill}}lc@{}}
        \toprule
        \textbf{Config} & \textbf{Value} \\
        \midrule
        Retrieval head ratio & 15\% \\
        Sliding window size & 8192 \\
        Sink tokens & 4 \\
        Low-dim size & 16 \\
        Top-$p$ & 0.9 \\
        Kernel block & 64 \\
        \bottomrule
    \end{tabular*}
    \vspace{-1.0em}
\end{wraptable}
\noindent\textbf{Settings.} Table~\ref{tab:exp_settings} summarizes the main configuration of \name. 
 We also conduct ablation studies on several key design settings of \name (see Appendix~\ref{appd:ablation}). For training, we use FineWeb~\citep{fineweb} and Dolma 3 Longmimo Mix~\citep{olmo2026olmo3}, from which we sample documents with lengths between 32K and 80K tokens. In the first stage, we train the low-dimensional projection parameters. In the second stage, we perform end-to-end training on corpora with an average length of 48K for about only 600 steps. The detailed training procedures are provided in Appendix~\ref{appd:training}.

\noindent\textbf{Baselines.} We compare \name against five representative sparse-attention baselines: RazorAttn~\citep{razorattn}, Minference~\citep{minference}, FlexPrefill~\citep{flexprefill}, Quest~\citep{Quest}, and SnapKV~\citep{snapkv}. For each method, we align the evaluation setting with both its official configuration and our own setup as much as possible to ensure a fair comparison. In particular, for RazorAttn we use the same 15\% retrieval-head ratio as in \name. For FlexPrefill, we set the cumulative-attention threshold $\gamma$ to 0.9 to match our top-$p$ threshold. For Quest, we strictly follow the official implementation, and do not apply sparse attention to the first two layers. Furthermore, to explicitly isolate and evaluate the advantage of our dynamic token budget, we implement a custom baseline that uses a static top-$k$ selection strategy, with $k$ empirically set to 4096.

\begin{figure}[t]
    \centering
    \includegraphics[width=\linewidth]{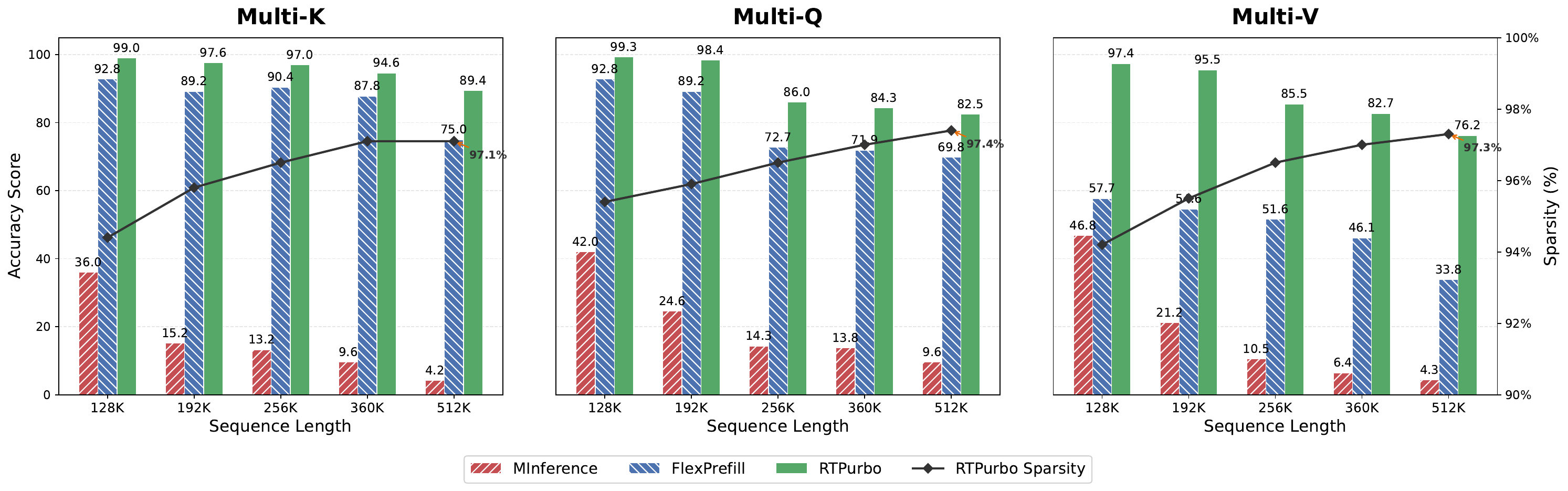}
    \caption{Accuracy and sparsity on ultra-long multi-hop tasks (128K--512K). Unlike baselines that collapse at extreme lengths, \name sustains robust accuracy while achieving high sparsity.}
    \label{fig:multi_benchmark}
\end{figure}

\noindent\textbf{Results on LongBench and RULER.} Table~\ref{tab:longbench} and Table~\ref{tab:ruler} summarize the evaluation results. Methods estimating global attention via recent queries (Minference, SnapKV) degrade significantly on multi-hop tasks (e.g., multi-Q and multi-K) where local context diverges from the full sequence. Similarly, the reliance on adjacent blocks of FlexPrefill causes severe drops on dispersed-evidence tasks like multi-V, while coarse block-level sparsity of Quest yields a general accuracy loss. As a training-free approach, RazorAttn also struggles on retrieval-heavy tasks (e.g., HotpotQA, Musique). Crucially, on long-context benchmarks such as RULER 64K, the fixed-budget variant performs poorly because it recalls too few tokens to preserve sufficient attention mass (see Appendix~\ref{appd:head_sparsity_diversity}). Furthermore, we extend our evaluation to ultra-long contexts (up to 512K). As illustrated in Figure~\ref{fig:multi_benchmark}, while baselines experience catastrophic degradation at extreme lengths, \name robustly sustains high accuracy. These comparisons confirm that \name with dynamic top-$p$ selection effectively adapts to varying query complexities, providing a trainable, fine-grained thresholding solution that strictly preserves accuracy. 
\begin{table}
\caption{Accuracy comparison on LongBench. The best average result among sparse-attention methods is shown in bold, and the second-best average result is underlined. Full attention is excluded from this ranking.}
\label{tab:longbench}
\centering
\newcommand{\rot}[1]{\rotatebox{60}{#1}}
\resizebox{\textwidth}{!}{%
\begin{tabular}{lccccccccccccccccc}
\toprule
\multirow{1}{*}[0.35cm]{\textbf{Longbench}} & \rot{2wiki} &
\rot{hotpot} & \rot{musique} & \rot{multi-en} & \rot{multi-zh} & \rot{qasper} & \rot{g-report} &
\rot{qmsum} & \rot{vcsum} & \rot{triviaqa} & \rot{trec} & \rot{lsht} & \rot{lcc} & \rot{repo-p} & \rot{PR-zh} & \rot{PR-en} &
\multirow{1}{*}[0.35cm]{\textbf{Avg. (\%)}} \\
\midrule
Full Attn & 42.08 & 54.64 & 38.30 & 52.89 & 65.99 & 39.25 & 31.89 & 23.77 & 13.55 & 89.93 & 85.50 & 61.50 & 35.08 & 27.61 & 99.75 & 99.00 & 53.80\\
RazorAttn & 40.31 & 50.92 & 29.15 & 53.04 & 65.90 & 39.72 & 31.61 & 23.59 & 13.85 & 89.26 & 85.00 & 59.50 & 35.01 & 32.05 & 99.00 & 99.75 & 52.98\\
Minference & 40.22 & 52.21 & 32.12 & 52.09 & 64.92 & 38.43 & 31.77 & 23.60 & 13.42 & 51.95 & 82.50 & 32.25 & 34.05 & 25.76 & 99.00 & 100 & 48.39\\
FlexPrefill & 36.48 & 52.76 & 35.71 & 51.69 & 65.86 & 37.44 & 31.63 & 24.00 & 13.10 & 90.59 & 80.50 & 50.25 & 32.85 & 24.93 & 93.50 & 69.50 & 49.42\\
Quest & 39.11 & 53.27 & 31.64 & 48.22 & 62.80 & 38.27 & 30.18 & 23.55 & 13.24 & 79.88 & 81.00 & 59.00 & 35.80 & 24.31 & 98.83 & 91.92 & 50.69\\
SnapKV & 42.56 & 53.10 & 38.14 & 52.70 & 65.50 & 39.30 & 16.11 & 17.68 & 11.03 & 82.03 & 77.00 & 57.21 & 35.00 & 25.76 & 99.00 & 99.75 & 50.74\\
\addlinespace[3pt]
\arrayrulecolor{black!60}\cdashline{1-18}[0.9pt/2pt]\arrayrulecolor{black}
\noalign{\vskip 3pt}
\multicolumn{18}{l}{\textbf{RTPurbo}}\\[1pt]
\shortstack[l]{\hspace{0.8em}w/ top-k} & 42.56 & 53.10 & 36.46 & 51.75 & 64.97 & 40.43 & 31.76 & 24.41 & 13.88 & 89.10 & 83.00 & 55.50 & 34.41 & 33.47 & 99.00 & 99.00 & \underline{53.30}\\
\shortstack[l]{\hspace{0.8em}w/ top-$p$} & 44.49 & 54.74 & 35.29 & 53.34 & 65.56 & 40.22 & 31.92 & 24.60 & 14.00 & 90.85 & 84.50 & 60.00 & 34.56 & 34.08 & 99.75 & 100 & \SOTA{54.24}\\
\bottomrule
\end{tabular}%
}
\end{table}
\begin{table*}[!t]
\caption{Accuracy comparison on RULER. The best average results among sparse-attention methods are shown in bold, and the second-best average results are underlined. Full attention is excluded from this ranking.}
\label{tab:ruler}
\centering
\scriptsize
\setlength{\tabcolsep}{7pt}
\renewcommand{\arraystretch}{1.1}
\begin{tabular}{lcccccccccc}
\toprule
\textbf{Method} &
\textbf{CWE} &
\textbf{FWE} &
\textbf{VT} &
\textbf{HotPot} &
\textbf{Squad} &
\textbf{multi-Q} &
\textbf{multi-V} &
\textbf{multi-K} &
\textbf{niah-S} &
\textbf{Avg.} \\
\midrule
\multicolumn{11}{c}{\textit{32K}} \\
\midrule
Full Attn & 81.80 & 91.47 & 98.00 & 64.60 & 72.63 & 99.95 & 98.85 & 99.66 & 99.93 & 89.65\\
RazorAttn & 82.50 & 90.07 & 92.04 & 65.00 & 71.88 & 99.60 & 97.65 & 99.66 & 99.80 & \underline{88.69}\\
Minference & 80.88 & 85.33 & 95.84 & 62.00 & 69.88 & 97.45 & 91.35 & 70.73 & 98.80 & 83.58\\
FlexPrefill & 75.26 & 90.00 & 96.56 & 60.40 & 71.18 & 98.20 & 60.30 & 98.66 & 100 & 83.40\\
Quest & 57.26 & 82.60 & 97.48 & 56.80 & 68.73 & 85.60 & 85.20 & 86.80 & 90.27 & 78.97\\
SnapKV & 76.23 & 89.92 & 94.23 & 58.92 & 69.20 & 94.00 & 71.10 & 97.46 & 99.86 & 83.43\\
\addlinespace[3pt]
\arrayrulecolor{black!60}\cdashline{1-11}[0.9pt/2pt]\arrayrulecolor{black}
\noalign{\vskip 3pt}
\multicolumn{11}{l}{\textbf{RTPurbo}}\\[1pt]
\shortstack[l]{\hspace{0.8em}w/ top-k} & 82.50 & 91.00 & 97.68 & 58.40 & 71.78 & 99.70 & 98.55 & 65.53 & 94.06 & 84.36\\
\shortstack[l]{\hspace{0.8em}w/ top-$p$} & 85.44 & 91.33 & 97.84 & 64.60 & 72.63 & 99.95 & 99.15 & 99.60 & 100 & \SOTA{90.06}\\
\midrule
\multicolumn{11}{c}{\textit{64K}} \\
\midrule
Full Attn & 65.28 & 84.00 & 96.80 & 63.40 & 69.83 & 99.50 & 97.60 & 99.66 & 100 & 86.23\\
RazorAttn & 65.96 & 83.80 & 91.52 & 62.80 & 69.33 & 98.95 & 95.20 & 98.73 & 99.73 & \underline{85.11}\\
Minference & 61.84 & 82.73 & 83.48 & 42.80 & 29.67 & 82.25 & 81.10 & 40.07 & 86.53 & 65.61\\
FlexPrefill & 51.04 & 80.07 & 92.72 & 58.00 & 68.63 & 96.35 & 55.50 & 97.60 & 100 & 77.77\\
Quest & 36.56 & 63.80 & 94.80 & 54.60 & 62.57 & 80.40 & 75.90 & 78.86 & 87.93 & 70.60\\
SnapKV & 60.08 & 72.00 & 90.00 & 47.70 & 68.15 & 91.15 & 60.50 & 94.00 & 98.73 & 75.81\\
\addlinespace[3pt]
\arrayrulecolor{black!60}\cdashline{1-11}[0.9pt/2pt]\arrayrulecolor{black}
\noalign{\vskip 3pt}
\multicolumn{11}{l}{\textbf{RTPurbo}}\\[1pt]
\shortstack[l]{\hspace{0.8em}w/ top-k} & 59.92 & 62.00 & 69.60 & 56.00 & 64.27 & 98.15 & 97.60 & 50.66 & 76.53 & 70.53\\
\shortstack[l]{\hspace{0.8em}w/ top-$p$} & 65.10 & 81.40 & 94.60 & 62.60 & 70.03 & 99.65 & 97.50 & 98.60 & 99.93 & \SOTA{85.49}\\
\bottomrule
\end{tabular}
\end{table*}

\begin{table}[t]
    \centering
    \scriptsize
    \setlength{\tabcolsep}{3.5pt}
    \renewcommand{\arraystretch}{0.96}
    \caption{Accuracy comparison on reasoning benchmarks. The best results among sparse-attention methods are shown in bold, and the second-best results are underlined. Full attention is excluded from this ranking.}
    \label{tab:reasoning}
    \resizebox{\linewidth}{!}{%
    \begin{tabular}{lccccccccc}
        \toprule
        \textbf{Method} & \textbf{AIME24} & \textbf{AIME25} & \textbf{MMLU-Bio} & \textbf{MMLU-Bus} & \textbf{MMLU-Chem} & \textbf{MMLU-CS} & \textbf{MMLU-Math} & \textbf{MMLU-Phi} & \textbf{MMLU-Phy}\\
        \midrule
        Full Attn & 86.67 & 86.67 & 89.20 & 87.40 & 88.00 & 86.10 & 93.60 & 69.94 & 90.80\\
        Quest & 46.67 & 46.67 & 88.28 & 84.28 & \underline{84.40} & \underline{82.68} & \underline{93.20} & \underline{69.34} & 87.70\\
        SnapKV & 43.33 & 46.67 & 88.14 & 81.27 & 76.32 & 80.49 & 90.00 & 66.40 & 84.42\\
        \addlinespace[3pt]
        \arrayrulecolor{black!60}\cdashline{1-10}[0.9pt/2pt]\arrayrulecolor{black}
        \noalign{\vskip 3pt}
        \multicolumn{10}{l}{\textbf{RTPurbo}}\\[1pt]
        \shortstack[l]{\hspace{0.8em}w/ top-k} & \underline{80.00} & \underline{80.00} & \underline{88.80} & \underline{86.60} & 51.40 & 50.49 & 71.60 & 51.70 & \underline{90.60}\\
        \shortstack[l]{\hspace{0.8em}w/ top-$p$} & \SOTA{86.67} & \SOTA{86.67} & \SOTA{89.80} & \SOTA{87.20} & \SOTA{87.80} & \SOTA{85.12} & \SOTA{93.60} & \SOTA{69.54} & \SOTA{91.00} \\
        \bottomrule
    \end{tabular}%
    }
\end{table} 
\noindent\textbf{Results on Reasoning Tasks.} Table~\ref{tab:reasoning} summarizes the reasoning benchmark results. These tasks exhibit extreme prompt-generation asymmetry: inputs are extremely short ($<$300 tokens), but generated reasoning traces are massively long (up to 32K tokens on AIME and averaging 10K on MMLU-PRO), shifting the bottleneck entirely to the decoding phase. As shown, \name with dynamic top-$p$ preserves near-lossless accuracy, perfectly matching the dense baseline on AIME (86.67). On the MMLU-PRO subcategories, it also remains consistently close to the full-attention baseline across all reported subjects, indicating that the sparse model retains strong general reasoning ability even under long decoding.

\subsection{Efficiency Evaluation}
\label{sec:efficiency_eval}

\begin{figure}[t]
    \centering
    \includegraphics[width=\linewidth]{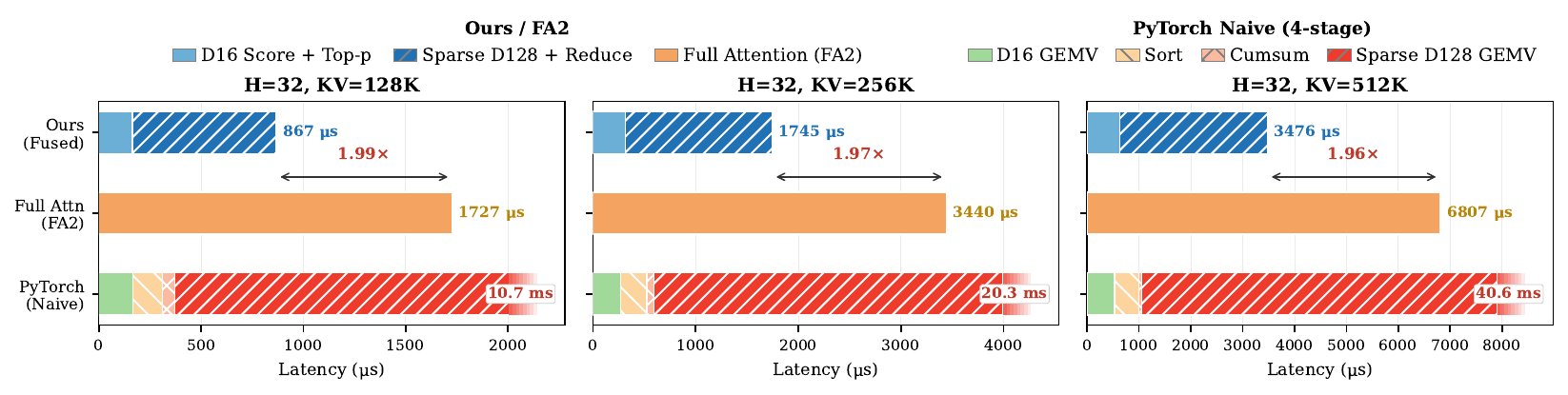}
    \caption{Sparse decoding speedup of \name.}
    \label{fig:sparse_decode_speedup}
\end{figure}


\noindent\textbf{Sparsity Analysis.} 
During prefill, our sparsity is deterministic: 15\% retrieval heads ($\mathcal{H}_{ret}$) perform dense attention, while 85\% local heads ($\mathcal{H}_{loc}$) attend only to 4 sink tokens and an 8192-token window. During decode, \name applies dynamic top-$p$ selection. Table~\ref{tab:dynamic_sparsity} profiles Layer 25 of Qwen3-Coder-30B-A3B, demonstrating that the optimal token budget is highly query-dependent. At 32K, \name retains just 468.8 active tokens for niah-S but dynamically expands to 2462.1 for multi-K. This 5$\times$ variance exposes the inherent flaw of rigid static top-$k$ methods, which inevitably suffer recall failure on complex queries or waste computation on simple ones. By adapting on the fly, we maintain high attention mass ($>$0.93) with exceptional sparsity (up to 89.2\% at 64K).

\begin{table}[t]
\centering
\small
\caption{Decode-stage dynamic sparsity of our top-$p$ mechanism. The active token budget adaptively scales with task complexity and context length. \textit{Compute/Memory Sparsity} follow the definitions in Section~\ref{sec:adaptive_sparse_attention}, while \textit{Active Tokens} and \textit{Attention Mass} denote the dynamically retained KV pairs per retrieval head and their preserved cumulative probability.}
\label{tab:dynamic_sparsity}
\begin{tabular}{clcccc}
\toprule
\textbf{Context} & \textbf{Task} & \textbf{Compute Sparsity} & \textbf{Memory Sparsity} & \textbf{Active Tokens} & \textbf{Attention Mass} \\
\midrule
\multirow{2}{*}{32K} & niah-S & 78.7\% & 76.2\% & 468.8 & $>$0.95 \\
                     & multi-K & 77.8\% & 74.4\% & 2462.1 & $>$0.96 \\
\midrule
\multirow{2}{*}{64K} & niah-S & 89.2\% & 87.7\% & 1126.8 & $>$0.93 \\
                     & multi-K  & 88.7\% & 85.2\% & 3316.1 & $>$0.94 \\
\bottomrule
\end{tabular}
\end{table}

Furthermore, we extend this decode-stage analysis to ultra-long contexts. As shown in Figure~\ref{fig:multi_benchmark}, \name robustly sustains high accuracy while its dynamic thresholding pushes sparsity to over 97.1\%\footnote{The reported ultra-long context sparsity is calculated as the average compute sparsity across all query heads; actual sparsity levels naturally vary among individual heads. We provide a more detailed analysis of per-head behaviors in Appendix~\ref{appd:headwise}.} at 512K. This confirms that our query-aware mechanism seamlessly extrapolates to extreme context lengths, maximizing efficiency without sacrificing recall. 



\noindent\textbf{Runtime Analysis.} We measure the speedup of a single attention layer under our sparse execution scheme. The results are shown in the left panel of Figure~\ref{fig:overall}. In the prefill phase, \name delivers substantial acceleration over FlashAttention-2 (FA2)~\citep{dao2024flashattention} across all tested context lengths, with speedups increasing from $2.83\times$ at 32K to $9.36\times$ at 1M. It also consistently outperforms the other sparse baselines in long context prefill. In the decode phase, \name also achieves stable speedups over FA2, improving from about $1.47\times$ at 32K to $2.01\times$ at 1M. 

In addition, we benchmark our single-operator top-$p$ decode kernel against both FA2 and a native PyTorch implementation. As shown in Figure~\ref{fig:sparse_decode_speedup}, our implementation consistently outperforms both baselines, confirming the efficiency of our specially designed top-p decode kernel.

\section{Conclusion}

We show that full-attention LLMs are already intrinsically sparse and can be transformed into efficient sparse inference systems with only minimal adaptation. Based on this view, we propose \name, a head-wise sparse attention framework built on retrieval/streaming head specialization, low-dimensional retrieval indexing, and dynamic top-$p$ selection.

Empirically, \name preserves near-lossless accuracy on both long-context and reasoning tasks while delivering substantial prefill and decode speedups. More broadly, our results suggest that native sparse pretraining is not the only path to efficient long-context inference: full-attention models can already support effective sparse execution with lightweight post hoc adaptation.

\section{Related work}

\noindent\textbf{Block-Sparse Attention.} Block-sparse methods reduce long-context cost by selecting only a subset of key--value blocks. QUEST~\citep{Quest} uses query-aware page ranking based on min--max key statistics, while MoBA~\citep{MoBA} treats sparse attention as block-level routing. BLASST~\citep{blasst}, SpargeAttention~\citep{spargeattn}, and Prism~\citep{prism} further improve block selection using softmax-contribution estimates, training-free refinement, or spectral criteria. These methods mainly differ in how they estimate block importance at the block level.

\noindent\textbf{Token-wise Attention.} Token-wise sparse attention first estimates token relevance and then applies exact attention only to the retained tokens. DSA~\citep{dsa} uses a lightweight learned indexer before top-$k$ selection, FASA~\citep{fasa} exploits the frequency structure induced by RoPE, and SnapKV~\citep{snapkv} compresses the KV cache using relevance to recent local queries.

\noindent\textbf{Pattern-Based Sparsity.} Another line of work adapts sparsity to head behavior. MInference~\citep{minference} assigns each head an offline-discovered sparse pattern, while FlexPrefill~\citep{flexprefill} makes the pattern selection context-aware. DuoAttention~\citep{duoattn} and RazorAttention~\citep{razorattn} instead partition heads into retrieval and streaming groups and treat them differently. Our method is most closely related to this line, but further introduces low-dimensional token indexing and dynamic top-$p$ selection for retrieval heads.

\newpage
{
\small
\bibliographystyle{plainnat}
\bibliography{references}
}

\newpage
\beginappendix

\section{Headwise Analysis of Local/Retrieval Patterns and Retrieval Sparsity}
\label{appd:headwise}

In this section, we analyze the distribution and properties of different head patterns in the two models used in our experiments, namely Qwen3-Coder-30B-A3B and Qwen3-30B-A3B-Think, to better understand their multi-head attention behaviors.

\subsection{Headwise Distribution.}
\label{appd:distribution}

In this section, we analyze the distribution of retrieval scores across all query heads in Qwen3-Coder-30B-A3B and Qwen3-30B-A3B-Think.

\begin{figure}[t]
    \centering
    \includegraphics[width=\linewidth]{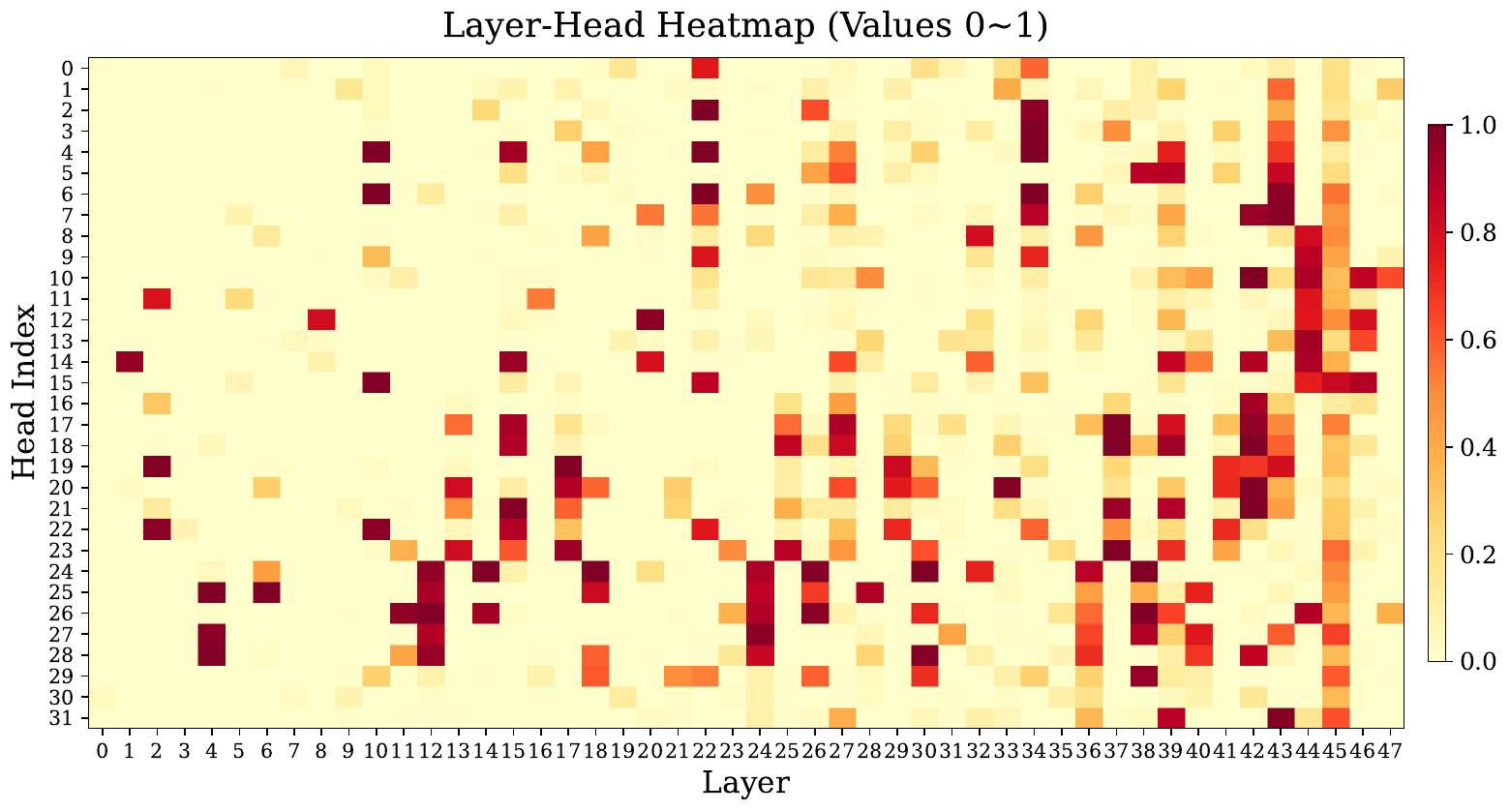}
    \caption{Head-wise retrieval scores of all query heads in Qwen3-Coder-30B-A3B. A higher score indicates stronger long-range retrieval ability.}
    \label{fig:headwise_attnmap}
\end{figure}

\noindent\textbf{Retrieval score distribution.}
Figure~\ref{fig:headwise_attnmap} presents the per-head retrieval score heatmaps for both models. We make the following observations:

\begin{itemize}[leftmargin=*, itemindent=0pt]
    \item \emph{Heads with strong retrieval ability are relatively few.} Most query heads receive only small retrieval scores, indicating limited long-range recall ability, while only a relatively small subset of heads consistently exhibits strong retrieval behavior. This suggests that the capacity for long-range token recall is concentrated in a minority of heads, whereas the majority primarily rely on local context or sink tokens for information processing.

    \item \emph{Retrieval heads concentrate in later layers.} High-scoring retrieval heads are distributed highly unevenly across layers, appearing almost exclusively in the latter half of the model. This is consistent with the layerwise computation pattern of LLMs: early layers are primarily responsible for local contextualization, where token representations are still evolving; later layers produce more stable, semantically rich representations that provide the necessary foundation for reliable long-range token retrieval~\citep{ghandeharioun2024patchscopes}.

    \item \emph{Head behavior is highly stable and input-agnostic.} The retrieval scores of individual heads remain highly consistent across different input documents, confirming our assumption in Section~\ref{sec:offline_calibration} that running offline calibration on a single long sequence is sufficient to robustly partition all query heads into retrieval and local sets.
\end{itemize}

\paragraph{Comparison across models.}
Qwen3-30B-A3B-Think, as a reasoning-specialized model, exhibits a head distribution pattern largely consistent with Qwen3-Coder-30B-A3B: only a relatively small subset of heads shows strong retrieval ability, and these heads are likewise concentrated in the later layers. This suggests that head specialization is a general intrinsic property of pretrained LLMs, rather than an artifact of a specific task or model architecture.

\subsection{Diversity of Sparsity between Retrieval Heads.}
\label{appd:head_sparsity_diversity}

While Section~\ref{appd:distribution} characterizes the binary distinction between 
retrieval and local heads, we further analyze the token-level sparsity patterns 
within the retrieval head set $\mathcal{H}_\text{ret}$. Table~\ref{tab:appendix_headwise_topp_tokens} 
reports the number of tokens recalled by top-$p$ ($p = 0.9$) for three representative 
retrieval heads under different sequence lengths. The results reveal substantial diversity 
in sparsity behavior across individual retrieval heads:

\begin{itemize}[leftmargin=*, itemindent=0pt]
    \item \emph{Retrieval heads differ drastically in their token budgets.} 
    Even under the same top-$p$ threshold and the same input sequence, different 
    retrieval heads retain very different numbers of active tokens. For instance, 
    at 64K context length, \texttt{L43H31} retains only 21 tokens to cover 90\% 
    of attention mass, while \texttt{L24H25} requires 24{,}621 tokens to reach 
    the same coverage threshold---a gap of over three orders of magnitude. This 
    confirms that a single fixed top-$k$ budget cannot simultaneously serve all 
    retrieval heads: it would either over-retain tokens for highly concentrated 
    heads or under-retain for diffuse ones.

    \item \emph{Head-level sparsity patterns are consistent across context lengths.}
    Despite the absolute token counts scaling with sequence length, the relative 
    ordering of heads by sparsity remains stable. \texttt{L43H31} consistently 
    retains far fewer tokens than \texttt{L2H19} and \texttt{L24H25} across all 
    tested lengths (32K, 64K, and 128K), suggesting that each head's tendency 
    toward concentrated or diffuse retrieval is an intrinsic property of that head 
    rather than a transient artifact of a specific input.

    \item \emph{Sparsity diversity motivates per-head dynamic thresholding.}
    The stark contrast between heads such as \texttt{L43H31} (highly concentrated) 
    and \texttt{L24H25} (broadly diffuse) highlights why our top-$p$ mechanism 
    is essential. By applying an independent, query-aware threshold to each head, 
    \name naturally accommodates this diversity: concentrated heads are served 
    with minimal token budgets, while diffuse heads expand their active sets only 
    when necessary. This per-head adaptivity is precisely what static top-$k$ 
    methods fundamentally cannot provide.
\end{itemize}

\begin{table}[ht]
    \centering
    \small
    \caption{Number of recalled tokens selected by top-$p$ ($p=0.9$) for representative retrieval heads under different sequence lengths.}
    \label{tab:appendix_headwise_topp_tokens}
    \begin{tabular}{lccc}
        \toprule
        \textbf{Retrieval Head} & \textbf{32K} & \textbf{64K} & \textbf{128K} \\
        \midrule
        L2 H19  & 4540 & 10468 & 9858 \\
        L24 H25 & 13836 & 24621 & 39614 \\
        L43 H31 & 55 & 21 & 42 \\
        \bottomrule
    \end{tabular}
\end{table}

\section{Ablations on \name Design Choices}
\label{appd:ablation}

We conduct ablation studies on several key design settings in \name.

\subsection{Ablation on Retrieval Head Ratio}

We compare two retrieval-head ratios in \name, namely 15\% and 30\%, on representative subcategories from MMLU-PRO and RULER.

\begin{table}[ht]
    \centering
    \small
    \caption{Ablation on retrieval-head ratio over MMLU-PRO subcategories.}
    \label{tab:appendix_mmlupro_ratio}
    \begin{tabular}{lcc}
        \toprule
        \textbf{MMLU-PRO Subcategory} & \textbf{\name (15\%)} & \textbf{\name (30\%)} \\
        \midrule
        Biology & 86.0 & 85.8 \\
        Computer Science & 76.8 & 76.8 \\
        History & 56.1 & 56.1 \\
        Math & 88.2 & 88.2 \\
        Philosophy & 59.1 & 59.0 \\
        \bottomrule
    \end{tabular}
\end{table}

\begin{table}[ht]
    \centering
    \small
    \caption{Ablation on retrieval-head ratio over RULER sub-benchmarks.}
    \label{tab:appendix_ruler_ratio}
    \begin{tabular}{lcc}
        \toprule
        \textbf{RULER 64K} & \textbf{\name (15\%)} & \textbf{\name (30\%)} \\
        \midrule
        FWE & 81.7 & 81.4 \\
        HotPot & 62.8 & 62.6 \\
        multi-Q & 99.7 & 99.7 \\
        multi-K & 98.8 & 98.6 \\
        niah-S & 99.9 & 99.9 \\
        \bottomrule
    \end{tabular}
\end{table}

\begin{table}[ht]
    \centering
    \normalsize
    \setlength{\tabcolsep}{6pt}
    \renewcommand{\arraystretch}{1.10}
    \caption{Benchmarks with 15\% vs. 10\% retrieval-head ratios.}
    \label{tab:appendix_ratio_small}
    \begin{tabular}{lcc}
        \toprule
        \textbf{Benchmark} & \textbf{15\%} & \textbf{10\%} \\
        \midrule
        MMLU-PRO Math & 88.2 & 79.3 \\
        MMLU-PRO CS & 76.8 & 70.2 \\
        RULER multi-K & 98.8 & 97.4 \\
        \bottomrule
    \end{tabular}
\end{table}

The results in Tables~\ref{tab:appendix_mmlupro_ratio},~\ref{tab:appendix_ruler_ratio}, and~\ref{tab:appendix_ratio_small} reveal a clear trade-off across different retrieval-head ratios. Increasing the ratio from 15\% to 30\% brings almost no accuracy improvement on either MMLU-PRO or RULER, while it directly reduces the overall sparsity and increases the training cost, since the first stage must optimize roughly twice as many low-dimensional projection parameters. In contrast, reducing the ratio further to 10\% causes a substantial accuracy drop on several representative benchmarks, indicating that the number of retrieval heads becomes insufficient to preserve robust long-range recall. Therefore, for the model used in our experiments, namely Qwen3-30B-A3B, a 15\% retrieval-head ratio provides the best balance between sparsity, training cost, and accuracy, and is thus the most practical design choice.

\subsection{Ablation on Low-dimension Size}
\label{appd:lowdim}

\begin{table}[ht]
    \centering
    \small
    \caption{Ablation on low-dimension size over representative benchmarks.}
    \label{tab:appendix_lowdim_accuracy}
    \begin{tabular}{lccc}
        \toprule
        \textbf{Benchmark} & \textbf{dim=4} & \textbf{dim=16} & \textbf{dim=32} \\
        \midrule
        MMLU-PRO Math & 89.1 & 88.2 & 88.2 \\
        MMLU-PRO CS & 76.9 & 76.8 & 76.8 \\
        RULER niah-S & 100 & 99.9 & 99.9 \\
        RULER HotPot & 63.0 & 62.8 & 62.7 \\
        \bottomrule
    \end{tabular}
\end{table}


\noindent\textbf{End-to-end Accuracy.} We compare different low-dimensional sizes in \name and study how the projection dimension affects both end-to-end task accuracy and the fitting quality of the low-dimensional relevance space. Specifically, we evaluate three representative settings, namely dim $=4$, dim $=16$, and dim $=32$, on a subset of representative benchmarks. Table~\ref{tab:appendix_lowdim_accuracy} summarizes the end-to-end accuracy results. We observe that dim $=16$ and dim $=32$ deliver nearly identical accuracy across all evaluated benchmarks. Since increasing the dimension from 16 to 32 doubles the number of trainable low-dimensional projection parameters, this result suggests that dim $=16$ is already sufficient to capture the attention distribution of retrieval heads accurately, and that further enlarging the projection space brings little practical benefit.

Interestingly, dim $=4$ yields the highest end-to-end accuracy in this comparison. However, this should not be interpreted as evidence that a smaller low-dimensional space provides a better approximation. Instead, Table~\ref{tab:appendix_l24h25_lowdim_topp_tokens} shows that dim $=4$ has substantially weaker fitting ability, which causes the top-$p$ selector to recall many more tokens in order to cover the same amount of attention mass. As a result, the actual sparsity becomes much lower under dim $=4$, and the corresponding accuracy improves because the sparse model behaves more similarly to a less aggressively pruned model.

\noindent\textbf{Fitting Ability of Low-dimension Space.} Table~\ref{tab:appendix_l24h25_lowdim_topp_tokens} reports the number of recalled tokens for retrieval head \texttt{L24H25} under different projection dimensions when using top-$p$ with $p=0.9$. The results show a clear trade-off between expressiveness and sparsity. When the projection dimension is too small, the compressed relevance space cannot faithfully model the full attention distribution, forcing the selector to retain substantially more tokens in order to recover the same amount of attention mass. This is exactly what happens at dimension 4, where the recalled token count rises sharply across all sequence lengths, indicating poor fitting ability and weak sparsity. Increasing the dimension to 16 dramatically improves the quality of the approximation and yields the smallest recalled-token budget overall, showing that this setting best captures the retrieval structure of the head while preserving high sparsity. Further increasing the dimension to 32 does not bring additional benefit; instead, it consistently requires more recalled tokens than dimension 16. This suggests that a moderately compact subspace is sufficient, and that a larger projection dimension may introduce unnecessary flexibility without improving token selection quality. Therefore, we choose dimension 16 as the default setting, since it achieves the best balance between fitting accuracy, sparsity, and computational efficiency.

\begin{table}[ht]
    \centering
    \small
    \caption{Number of recalled tokens for retrieval head \texttt{L24H25} under top-$p$ ($p=0.9$) with different input lengths and projection dimensions.}
    \label{tab:appendix_l24h25_lowdim_topp_tokens}
    \begin{tabular}{lccc}
        \toprule
        \textbf{Dimension} & \textbf{32K} & \textbf{64K} & \textbf{128K} \\
        \midrule
        4  & 25256 & 45280 & 89081 \\
        16 & 13771 & 25725 & 49133 \\
        32 & 15526 & 28464 & 54727 \\
        \bottomrule
    \end{tabular}
\end{table}


\section{Details of the Two-Stage Training Pipeline}
\label{appd:training}

In this section, we present the detailed procedure of our two-stage training pipeline and analyze the overall training workflow for Qwen3-Coder-30B-A3B.

\begin{table}[t]
    \centering
    \begin{minipage}[t]{0.49\linewidth}
        \centering
        \normalsize
        \setlength{\tabcolsep}{6pt}
        \renewcommand{\arraystretch}{1.10}
        \captionof{table}{Stage-1 training configuration.}
        \label{tab:appendix_stage1_settings}
        \begin{tabular}{lc}
            \toprule
            \multicolumn{2}{c}{\textbf{Training Settings}} \\
            \midrule
            max-lr & 1e-3 \\
            warmup-steps & 100 \\
            lr-scheduler & consine \\
            weight-decay & 0.01 \\
            max-grad-norm & 1.0 \\
            \bottomrule
        \end{tabular}
    \end{minipage}\hfill
    \begin{minipage}[t]{0.49\linewidth}
        \centering
        \normalsize
        \setlength{\tabcolsep}{6pt}
        \renewcommand{\arraystretch}{1.10}
        \captionof{table}{Stage-2 training configuration.}
        \label{tab:appendix_stage2_settings}
        \begin{tabular}{lc}
            \toprule
            \multicolumn{2}{c}{\textbf{Training Settings}} \\
            \midrule
            max-lr & 3e-6 \\
            warmup-steps & 200 \\
            lr-scheduler & constant \\
            weight-decay & 0.01 \\
            max-grad-norm & 1.0 \\
            global-batch-size & 8 \\
            micro-batch-size & 1 \\
            \bottomrule
        \end{tabular}
    \end{minipage}
\end{table}

\subsection{Details of Low-dimension Projection Training}
\label{appd:train_stage_1}

For Qwen3-Coder-30B-A3B, the total number of query heads is 1536, with head dimension $d_h=128$. After offline calibration, we select the top 210 heads with the highest retrieval scores as retrieval heads, corresponding to roughly 15\% of all heads. For stage-1 training, we construct the training set by sampling 8,000 sequences from the open-source FineWeb dataset, each with a length between 32K and 80K tokens. For each retrieval head, our objective is to align the original attention-score distribution with the compressed attention-score distribution, while training only the low-dimensional query/key projection parameters. Under our setting, each head introduces $2 \times 128 \times 16 = 4096$ trainable parameters, corresponding to the query and key projections with head dimension 128 and low dimension 16. Therefore, the total number of trainable parameters in stage 1 is $210 \times 4096 \approx 8.6 \times 10^5$, i.e., about 840K parameters in total.

\begin{figure}[ht]
    \centering
    \begin{subfigure}[t]{0.49\linewidth}
        \centering
        \includegraphics[width=\linewidth]{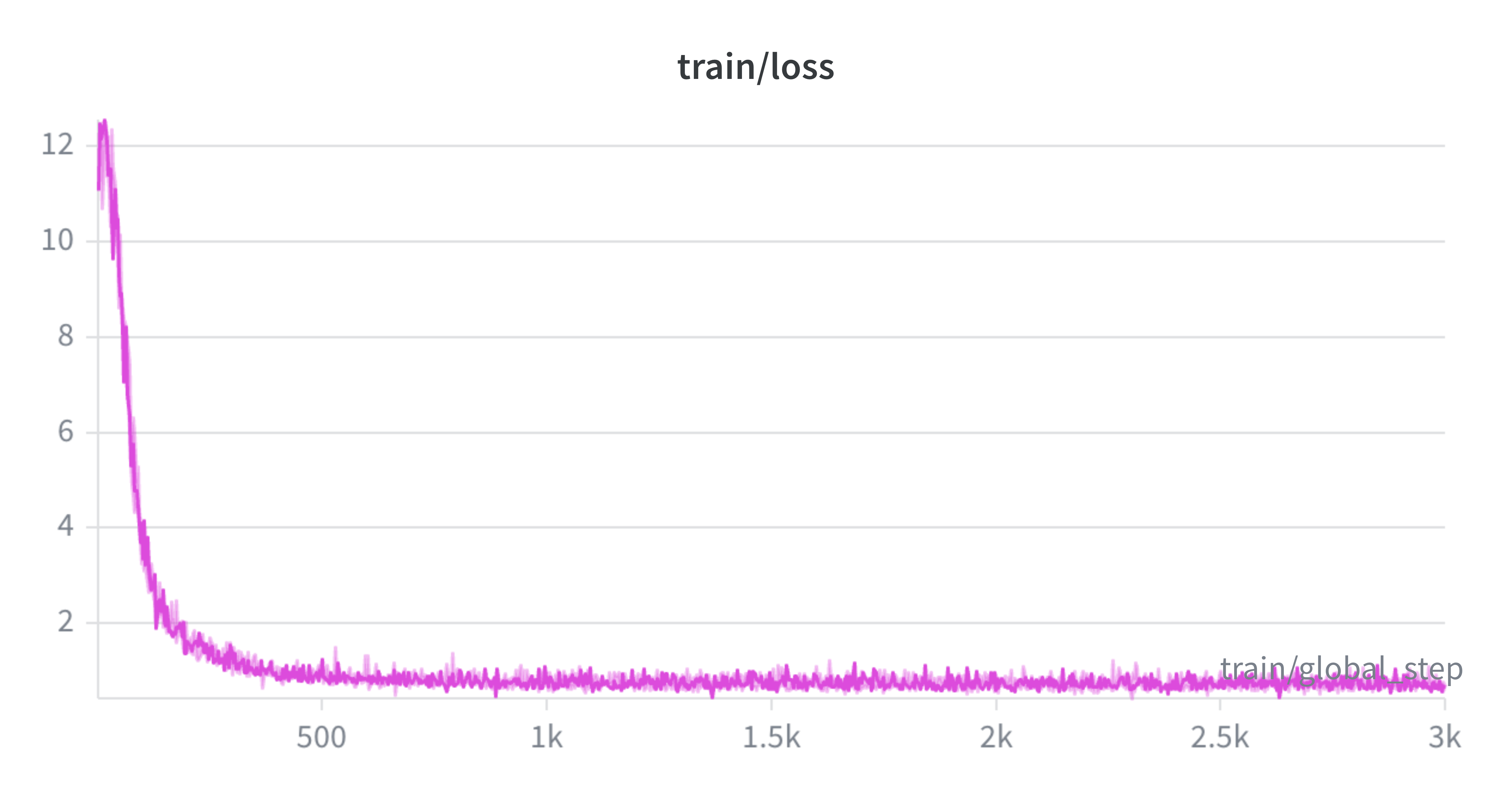}
        \caption{Training loss curve of retrieval head L24H25 during stage-1 low-dimension projection training.}
        \label{fig:stage1_indexer_loss}
    \end{subfigure}\hfill
    \begin{subfigure}[t]{0.49\linewidth}
        \centering
        \includegraphics[width=\linewidth]{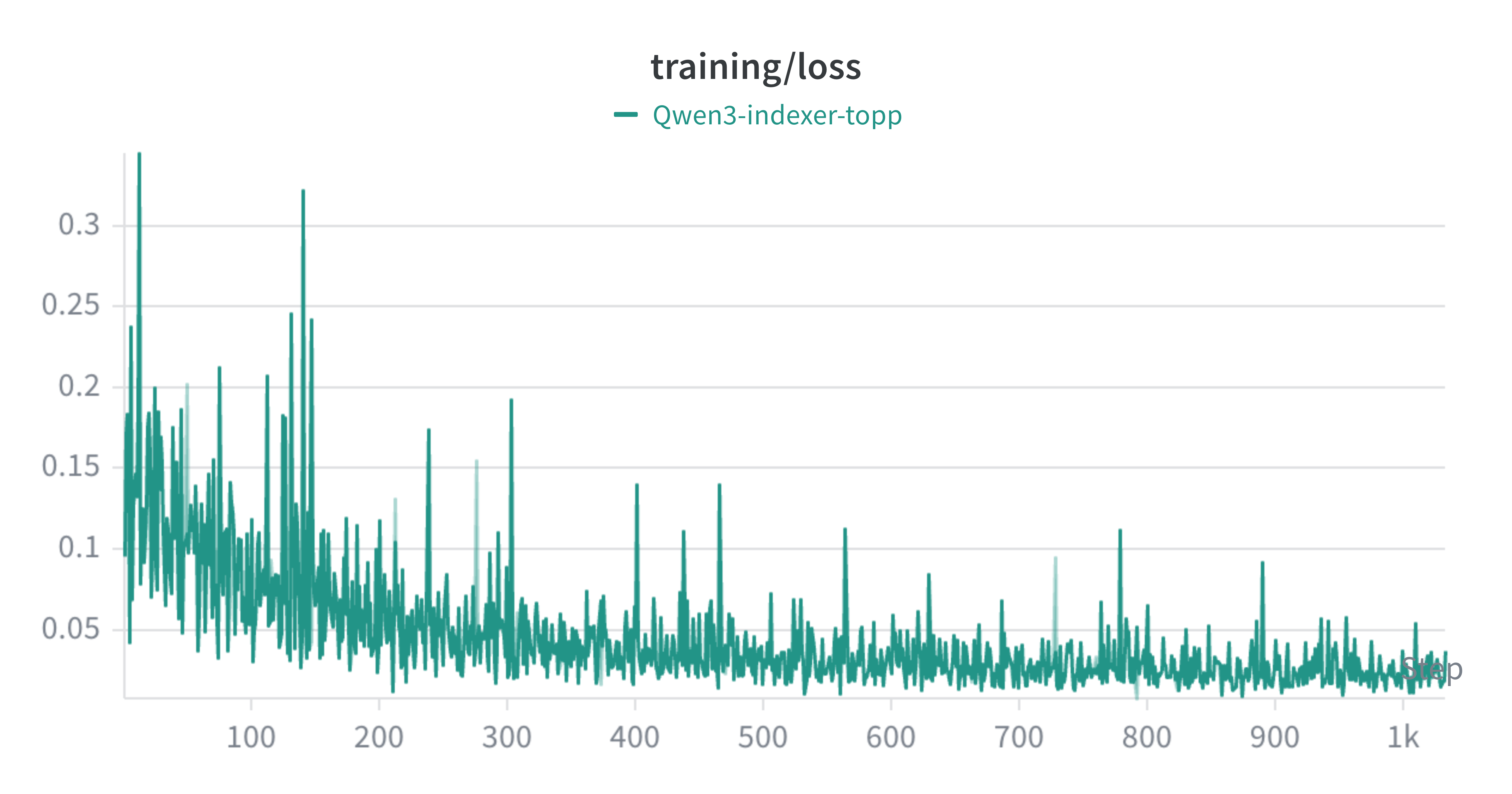}
        \caption{End-to-end training loss curve during stage-2 self-distillation.}
        \label{fig:stage2_end2end_loss}
    \end{subfigure}
    \caption{Training loss curves of the two-stage training pipeline.}
    \label{fig:two_stage_training_loss}
\end{figure}

The training settings are summarized in Table~\ref{tab:appendix_stage1_settings}. Specifically, the learning rate is linearly warmed up from 0 to the peak value of $10^{-3}$ over the first 100 steps, and then decayed with a cosine annealing schedule. As shown in Figure~\ref{fig:stage1_indexer_loss}, the training loss of Layer 24 Head 25 already converges well within about 600 steps, and we observe highly similar convergence behavior for the other retrieval heads. Since each training sequence contains 48K tokens on average, the total token budget of this stage is approximately $48\mathrm{K} \times 600 \simeq 30\mathrm{M}$ tokens.

\subsection{Details of End-to-End Self-distillation Training}
\label{appd:train_stage_2}

For stage-2 training, we sample 8,000 long reasoning examples in dialogue format from Dolma 3 Longmimo Mix. Each sequence is longer than 32K tokens, the average sequence length is about 48K tokens, and the average number of training label tokens is about 300. We first run a forward pass on the original model over the entire training set, extract the top-10 next-token prediction logits at each position, and cache them as distillation targets. During training, we attach the low-dimensional parameters learned in stage 1 to the selected retrieval heads and keep these parameters frozen, while performing end-to-end optimization over the model weights. The objective of this stage is to align the logits of the sparse model with those of the original model before sparsification.

The training settings are summarized in Table~\ref{tab:appendix_stage2_settings}. In particular, we use a relatively small learning rate to avoid drifting away from the model's original capabilities during end-to-end finetuning. As shown in Figure~\ref{fig:stage2_end2end_loss}, the training loss of stage 2 converges within about 600 steps. The full training corpus contains roughly 180M tokens in total, while the actual number of label tokens involved in learning is only about 1.2M.

\section{Limitation} \label{appd:limitation}

Although \name achieves strong efficiency--accuracy trade-offs, it still has several limitations.

\begin{itemize}[leftmargin=*, itemindent=0pt]
    \item \emph{Dependence on stable head specialization.} Our method relies on the empirical observation that attention heads can be partitioned into retrieval and local groups through offline calibration. While this behavior is stable in the models we study, the quality of this partition may degrade for models with weaker head specialization or under substantial domain shift.

    \item \emph{Incomplete sparsification and limited evaluation scope.} In the current design, retrieval heads still use full dense attention during prefill. In addition, our experiments mainly focus on the Qwen3 family and on long-context and reasoning workloads, so broader validation on other architectures and domains is still needed.
\end{itemize}

We expect these limitations to be addressed by future work on more adaptive head routing, stronger prefill sparsification, and broader cross-model evaluation.





\end{document}